\documentclass[a4paper]{article}
\usepackage{times}
\usepackage{amsmath,graphicx,subfigure,amssymb}
\usepackage{algorithmic, algorithm,color,flushend,spconf}
\usepackage{colortbl}
\usepackage{subfigure} 
\usepackage{setspace,ifthen,amsmath, amsthm, amssymb, graphicx,amsfonts, multirow} 
\newcommand{\CL}{\cellcolor[gray]{0.8}}

\usepackage{fancyheadings}
\title{Radon-Gabor Barcodes for Medical Image Retrieval}

\name{Mina Nouredanesh$^1$, H.R. Tizhoosh$^{2*}$, Ershad Banijamali$^3$,  James Tung$^1$} \vspace{0.05in}
\address{$^1$ Department of Mechanical and Mechatronics Engineering, University of Waterloo, ON, Canada\\
$^2$ KIMIA Lab, University of Waterloo, Canada, \emph{tizhoosh@uwaterloo.ca} [$^*$corresponding author]\\
$^3$ Cheriton School of Computer Science, University of Waterloo, ON, Canada}

\begin{document}

\maketitle
\begin{abstract}
In recent years, with the explosion of digital images on the Web, content-based retrieval has emerged as a significant research area. Shapes, textures, edges and segments may play a key role in describing the content of an image. Radon and Gabor transforms are both powerful techniques that have been widely studied to extract shape-texture-based information. The combined Radon-Gabor features may be more robust against scale/rotation variations, presence of noise, and illumination changes. The objective of this paper is to harness the potentials of both Gabor and Radon transforms in order to introduce expressive binary features, called barcodes, for image annotation/tagging tasks. We propose two different techniques: Gabor-of-Radon-Image Barcodes (GRIBCs), and  Guided-Radon-of-Gabor Barcodes (GRGBCs). For validation, we employ the IRMA x-ray dataset with 193 classes, containing 12,677 training images and 1,733 test images. A total error score as low as 322 and 330 were achieved for GRGBCs and GRIBCs, respectively. This corresponds to $\approx 81\%$ retrieval accuracy for the first hit. 
\end{abstract}



\section{Introduction}
Medical imaging plays a key role in modern healthcare resulting in emergence of massive image databases and picture archiving and communication systems (PACS). Due to the large quantity of available images, the need has grown for an efficient and accurate method to search in image databases. Searching based on descriptive text annotations exhibits several limitations. First, manual annotations of massive medical images by medical experts are time-consuming and expensive to implement. Moreover, since medical images contain several anatomical regions, the image content is difficult to be concretely described in words, i.e., irregular shapes of lesions cannot easily be expressed in textual form \cite{1}. 

Content-based image retrieval (CBIR) complements the conventional text-based retrieval of images by using visual features, i.e., color, texture, and shape, as search criteria \cite{2}.  One of the most important challenges in medical CBIR is the sheer number of images collected from different modalities, e.g., x-ray images that account for a large part of all medical images. Average-sized radiology departments produce several tera-bytes of data annually. The valuable information contained in images is difficult to capture, and generally stays unused after archiving. 

In recent years, there has been an increasing number of works proposing different methods for extraction of binary features, e.g., local binary patterns (LBP), Binary Robust Invariant Scalable Keypoints (BRISK) and Radon barcodes (RBC). Binary features have distinct benefits compared to conventional features. For instance, one million bag of word (BoW) features of 10,000 dimensions would need 1GB memory (with a compressed version of the inverted file). Conversely, for binary embeddings, the memory consumption is much lower (48MB for one million 48-bit binary codes) \cite{6}.
 
Recently, the concept of ``barcode annotation'' \cite{3} has been proposed  as a method for fast image search and content-based annotation. Barcodes can be embedded in the medical images files as supplementary information to increase the accuracy of image retrieval. Barcodes can also improve the performance of the conventional BoW methods. 

This paper aims to further develop the idea of ``content-based barcodes'' from a different point of view, namely by applying Gabor transform in conjunction with Radon transform, to generate Gabor-Radon-based barcodes for extraction of shape-texture information from medical images. Inspired by Radon barcodes (RBC), two distinct methods will be proposed in this paper: Guided-Radon of Gabor Barcodes (GRGBCs), and Gabor-of-Radon-Image Barcodes (GRIBCs).

\section{Motivation}

Radon and Gabor Transforms are both powerful techniques that have been widely explored in literature.  Several studies have attempted to modify Radon-based features to  generate features invariant to geometric variations (translation, rotation, scaling) \cite{13,14}. 
 According to \cite{13}, Radon transform mainly captures the shape-based details of the images. Due to the characteristics of the x-ray images, which mainly consists of two regions (a bright foreground region and a dark background), texture may be an appropriate feature for describing the contents of some body regions, e.g., breast tissue in mammograms \cite{17}. 

Texture analysis has been an active research field with numerous algorithms developed based on different models, e.g., grey-level co-occurrence (GLC) matrices and Markov random field (MRF) model \cite{4}. However, these well-known spatial domain texture analysis models usually analyze the image at a single scale. This can be improved by employing a spatial-frequency representation. In the past years, wavelets have attracted much attention for providing a multi-resolution analysis of the image. An important class of wavelets are Gabor filters, which are designed to mimic some functions of human vision, and have the capability of capturing the filtered-correlated responses at different scales and orientations. 

Gabor filters have been widely used to extract texture features from images for segmentation \cite{20,21}, object detection and biometric identification \cite{12}, and image retrieval \cite{8,9,10}. In \cite{5}, the authors have compared the performance of various texture classification methods, i.e., dyadic wavelet, wavelet frame, Gabor wavelet and steerable pyramids, and have observed that the Gabor-based methods provide superior performance compared to other texture methods. The most important advantage of Gabor filters is that they exhibit robustness against rotation, scale, and translation. Furthermore, they are robust against photometric disturbances, such as illumination changes and image noise \cite{11}. 

Based on the aforementioned reasons, using a combination of Gabor and Radon features is expected to provide more information regarding the image content, and consequently resulting in more discriminative features, and in higher retrieval accuracy.

\section{Background}
\vspace{-.075cm}
\textbf{Radon Barcodes --} Radon transform may be used to collect features from within an image. These features are the projections of the image intensity along radial lines oriented at a specific angle $\theta$. Considering the image $I$ as a function $f(x,y)$, one can project $f(x,y)$ along a number of projection angles $N_\theta$. The projection is basically the sum (integral) of $f(x,y)$ values along lines constituted by each angle $\theta$. The projection creates a new image $R(\rho,\theta)$ using the Dirac delta function  $\delta$ given as
\begin{equation}
R(\rho,\theta) = \int\limits_{-\infty}^\infty\!\int\limits_{-\infty}^\infty f(x,y)\delta(\rho\!-\!x\cos\theta\!-\!y\sin\theta) dx dy,
\end{equation}
where $\rho=x\cos\theta+y\sin\theta$. As we discussed earlier, according to \cite{3}, Radon Barcode (RBC) can be generated by thresholding all projections (lines) for individual angles to assemble a barcode of all thresholded projections.
The threshold is calculated by taking the median of all non-zero projections values.
%

\textbf{Gabor Transform --} In the spatial domain, a two-dimensional Gabor filter is a Gaussian kernel function modulated by a complex sinusoidal plane wave:
\begin{equation}
G_{(x,y)}\!=\!\frac{f^2}{\pi \gamma \eta} \exp\!{\left(\!-\frac{x'^2\!+\!\gamma^2 y'^2}{2\sigma^2}\!\right)}\!\exp{\left(j2\pi\! f\!x'\!+\!\phi\right)}
\end{equation}

Here $x'=x\cos\theta+y\sin\theta$, $y'=-x\sin\theta+y\cos\theta$, $f$ is the frequency of the sinusoid, $\theta$ indicates the orientation of the normal to the parallel stripes of a Gabor function, $\phi$ is the phase offset, $\sigma$ stands for the standard deviation of the Gaussian envelope, and $\gamma$ is the spatial aspect ratio which specifies the ellipticity of the support of the Gabor function \cite{12}. Given an image $I(x,y)$, the response of Gabor filter is the convolution of each Gabor window in the Gabor Filter Bank GFB$(N_u, N_v, s, t)$,  with image $I$, and is given by 
\begin{equation}
\psi_{(u,v)} (x,y)=\sum_s \sum_t I(x\!-\!s,y\!-\!t)*G_{(u,v)} (s,t),
\end{equation}       
where \textcolor{black}{$N_u$ and $N_v$ are the number of scales and orientations, respectively. Each $u \in \{1,2,...,N_u\}$ and $v \in \{1,2,...,N_v\} $ correspond to a specific Gabor window in GFB ($N_g=N_u \times N_v$ Gabor windows)}, and $s$, $t$ are the filter \textcolor{black}{window} size. The function $\psi_{(u,v)} (x,y)$ forms complex-valued function including real and imaginary parts. Other wavelets are dilated, shifted, and rotated versions of these values. In \cite{8} and \cite{12}, the authors presented an image retrieval and feature extraction method based on Gabor filter, in which texture features were extracted by calculating mean and variation of the Gabor filtered of natural images; however, the filter responses that result from the application of a filter bank of Gabor filters can be used directly as texture features.

\section{Proposed Barcodes}
\begin{figure}[tb]
\begin{center}
\vspace{0.05in}
\includegraphics[width=0.8\columnwidth]{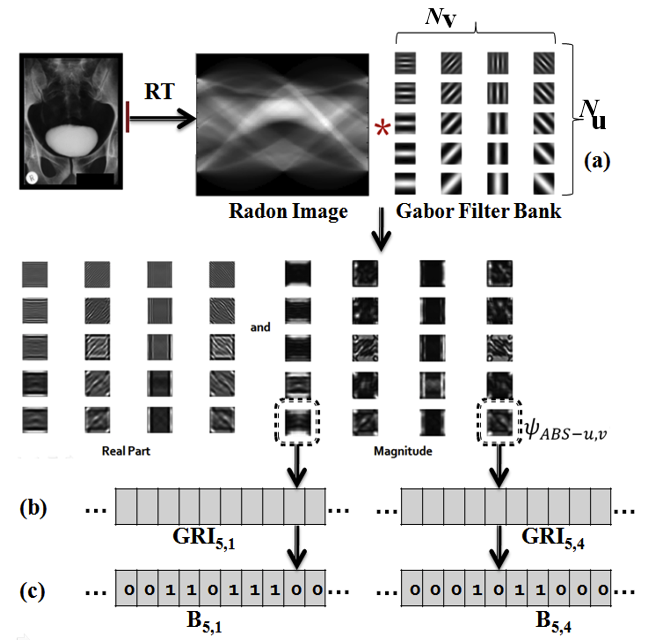}
\caption{Proposed GRIBCs algorithm (Algorithm \ref{alg:GaborRadon}).}
\label{fig:fig3}
\end{center}
\end{figure}

In this section we propose two methods: Gabor of Radon Barcodes (GRIBCs), and Radon-Gabor Barcodes (GRGBCs).
\\
\textbf{Gabor of Radon Barcodes (GRIBCs) --} The generation of Gabor of Radon Image Barcode for image $I_i$ (GRIBC$_i$) consists of three main steps: 1) Applying Radon transform on $I_i$ and obtaining the 2D Radon Image, 2) Applying Gabor filters on Radon image, and 3) Binarization of the feature vectors. In order to receive same-length feature vectors, all x-ray images should be resized into $R_N\!\times\! C_N$ images, i.e., $R_N\!=\!C_N\!=\!2^n$,  $n\!\in\! \mathbb{N}^+$. In this paper, to extract the Gabor of Radon image Barcodes, we use $R_N\!=\!C_N\!=\!128$. The resized images are then Radon transformed with 180 projections ($N_\theta\!=\! 180$), resulting in 2D matrices in Radon domain, that are called Radon-Image ($I_\textrm{Radon}$) \textcolor{black}{(Figure \ref{fig:fig3})} or shadowgrams \cite{19}. To extract the feature vectors with smaller dimensions, and to make the Barcodes more memory-efficient, the Radon images are resized into $M\!\times\! N\!=32\!\times\!32$ (considering the coordinates as $x$ and $y$ instead of $\rho$ and $\theta$). At this time, the Gabor filter with $u$ scales, $v$ directions, and filter window size $(s,t)$, (Gabor filter bank $GFB(N_u,N_v,s,t)$)  is applied to  $I_\textrm{Radon}$, resulting in $\psi_{(u,v,i)}(x,y)$. Since $\psi_{(u,v,i)}(x,y)$ consists of complex numbers, the magnitude or absolute value of each resulted $\psi_{(u,v,i)}$ is calculated ($\psi_{(ABS-u,v,i)}(x,y)$). Moreover, since the adjacent pixels in an image are usually highly correlated, we can reduce this information redundancy by downsampling the feature by a factor of $d1$, $d2$ for the column and row, respectively \cite{12} (we set $d1=d2=4$). The Gabor of Radon Image feature vector $GRI_{(u,v,i)}$ is obtained by transforming the downsampled $\psi_{(ABS-u,v,i)}$ to a vector.
\begin{algorithm}[t]
\caption{Gabor-of-Radon Image Barcodes }
\begin{algorithmic}[1]
\label{alg:GaborRadon}
\STATE Initialize GRIBC$_\textrm{i}\leftarrow \emptyset$
\FOR{all images $I_i$}
	\STATE $R_N\!=\!C_N\!\leftarrow 128$
	\STATE $I_i \leftarrow$ Normalize($I_i , R_N , C_N$)
	\STATE Set number of projection angles $N_\theta$
	\STATE $I_\textrm{Radon,i}\leftarrow$ RadonTransfom($I$)
	\STATE $I_\textrm{Radon} \leftarrow $ resize($I_\textrm{Radon,i}$, [32 , 32])
	\FOR{$\forall u$\textcolor{black}{$\in \{1,...,N_u\}$} and $\forall v$\textcolor{black}{$\in \{1,...,N_v\}$}}
		\STATE $\psi_{u,v}(x,y)\leftarrow$ Gabor($I_\textrm{Radon,i}$)
		\STATE $\psi_{(ABS-u,v)} (x,y)\leftarrow |(\psi_{(u,v)} (x,y)|$
	\textcolor{black}{\STATE Resample $\psi_{(ABS-u,v)}(x,y)$ with $d_1\!\times\!d_2$ coefficients}
		\STATE GRI$_{u,v,i} \leftarrow$ ReshapeToVector($\psi_{(ABS-u,v)}(x,y)$)
		\STATE Get threhsold: $T_{u,v,i}\leftarrow$ FindMedian($GRI_{u,v,i}$)
		\STATE Binarize: $B_{u,v,i}\leftarrow$   Find(GRI$_{u,v,i} \geq T_{u,v,i}$)
		\STATE Append barcode: \\GRIBC$_{i}\!\leftarrow\!$ AppendRow (GRIBC$_{i},B_{u,v,i}$)
	\ENDFOR
\ENDFOR
\end{algorithmic}
\end{algorithm}

\begin{figure}[t]
\begin{center}
\includegraphics[width=0.85\columnwidth]{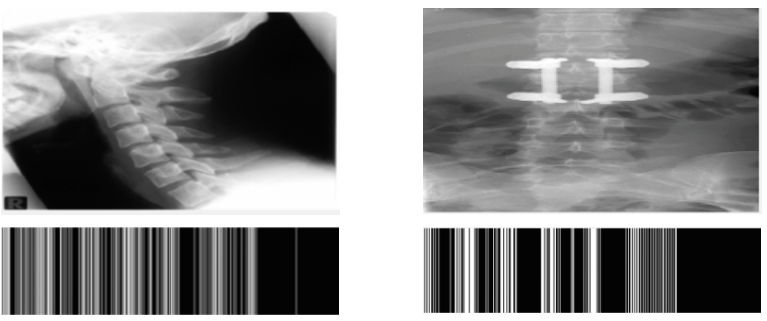}
\caption{Sample barcodes for IRMA images.}
\label{fig:fig4}
\end{center}
\vspace{-.8cm}
\end{figure}

To generate GRIBC$_i$ for the image $I_i$, the thresholds $(T_{u,v,i})$ are calculated for each of the GRI$_{(u,v,i)}$ vectors separately. Each threshold $T_{u,v,i}$ is employed to binarize the corresponding GRI$_{(u,v,i)}$ and to produce the binary vector $B_{(u,v,i)}$. The final GRIBC$_i$ for image $I_i$ is obtained by appending these binarized vectors GRIBC$_i= [B_{1,1}\cdots \textcolor{black}{B_{N_u,N_v}}]$. The details of the algorithm are provided in Algorithm  \ref{alg:GaborRadon}. Figure \ref{fig:fig3} visualizes the major steps. Sample GRI barcodes are displayed in Figure \ref{fig:fig4}. Generally, by applying ($N_g$) Gabor windows, the Gabor feature vector of a M by N image is a vector with length \textcolor{black}{$(M\!\times\!N\times \!N_g)/(d1\!\times\! d2$)} \cite{12} \textcolor{black}{(e.g., for $N_g\!=\!80$, downsampling coefficients of $d_1=d_2=4$, and $M\!=\!N\!=\!32$, the vector dimension is $\frac{81920}{4\times4}\!=\!5120$).}

\begin{figure}[!t]
\begin{center}
\includegraphics[width=0.9\columnwidth]{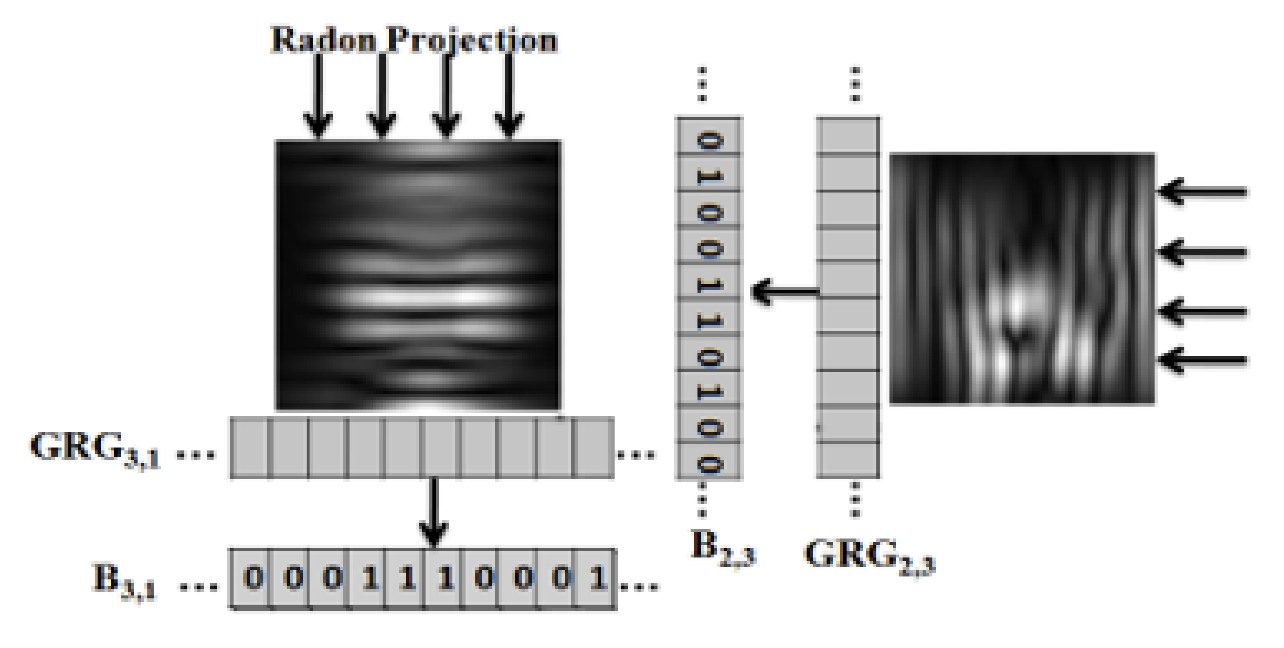}
\caption{In GRGBC algorithm, rays are applied perpendicularly to the direction of stripes in each Gabor filter in GFB.}
\label{fig:Guided}
\end{center}
\end{figure}

\begin{figure}[!t]
\begin{center}
\includegraphics[width=0.85\columnwidth]{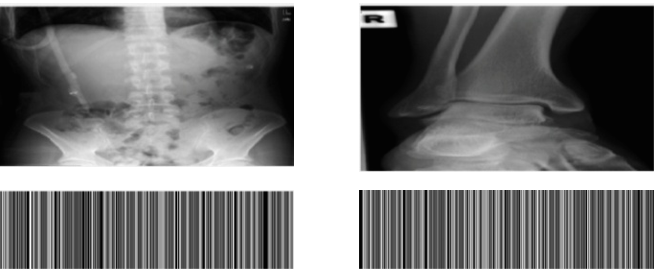}
\caption{Sample barcodes for IRMA images.}
\label{fig:fig6}
\end{center}
\vspace{-.7cm}
\end{figure}

\textbf{Guided Radon-Gabor Barcodes (GRGBCs) --} The Guided Radon of Gabor method is a new approach for extraction of shape-texture-based features. The word ``guided'' emphasizes that Radon projections are applied perpendicularly to the direction of the parallel stripes in each filter in the Gabor filter bank. In order to receive same-length feature vectors, all x-ray images should be resized into $R_N\!\times\! C_N$ images. We extracted the GRG barcodes from resized  $64\!\times\! 64$ images. Clearly, in this algorithm number of Radon projections is equal to the number of orientations in Gabor filter bank ($N_\theta\!=\! \textcolor{black}{N_v}$).
Generation of GRGBC$_i$ for the image $I_i$ comprises five  steps: 1) Obtaining \textcolor{black}{$N_g$} Gabor filtered responses by applying \textcolor{black}{GFB$(N_u,N_v,s,t)$} to the image $I_i$,  2) applying Radon transform to each $\psi_{(ABS-u,v)} (x,y)$ separately such that $\theta_{Radon}\leftarrow \theta_{Gabor}+90^\circ$ \textcolor{black}{($\theta_{Gabor}$ corresponds to a specific orientation $v$ in GFB)}, 3) transforming the 2D results into a row vector (resulting in Guided-Radon-of-Gabor feature vector for each $u$ and $v$ \textcolor{black}{(GRG$u,v,i$)}, 4) binarizing \textcolor{black}{GRG$_{u,v,i}$} (and producing $B_{u,v,i}$) by calculating the medians $T_{u,v,i}$ of each \textcolor{black}{GRG$_{u,v,i}$}  as a threshold, and 5) appending all $B_{u,v,i}$ for the image $I_i$ to generate the final barcode GRGBC$_i$ for that image. 
The details of the algorithm are provided in Algorithm \ref{alg:GuidedGaborRadon} and Figure \ref{fig:Guided}. The generated GRG barcodes are displayed for \textcolor{black}{two} sample images in Figure \ref{fig:fig6}.
\begin{algorithm}[!t]
\caption{Guided Radon-of-Gabor Barcodes}
\begin{algorithmic}[1]
\label{alg:GuidedGaborRadon}
\STATE Initialize GRGBC$_\textrm{i}\leftarrow \emptyset$, $v=8$ 
\STATE Set $\theta\!=\! \{0^\circ\!,22.5^\circ\!,45^\circ,67.5^\circ\!,90^\circ\!,112.5^\circ\!,135^\circ\!,157.5^\circ\!\}$
\FOR{all images $I_i$}
	\STATE $R_N\!=\!C_N\!\leftarrow 64$
	\STATE $I_i \leftarrow$ Normalize($I_i , R_N , C_N$)
	\FOR{$\forall u$\textcolor{black}{$\in \{1,...,N_u\}$} and $\forall v$\textcolor{black}{$\in \{1,...,N_v\}$}}
	\STATE $\psi_{(u,v,i)}(x,y)\!\leftarrow\!\sum\limits_s\sum\limits_t I_{i} (x\!-\!s,y\!-\!t)*G_{(u,v)} (s,t)$ 
	\STATE $\psi_{(ABS-u,v)} (x,y)\leftarrow |(\psi_{(u,v)} (x,y)|$
	\textcolor{black}{\STATE Resample $\psi_{(ABS-u,v)}(x,y)$ with $d_1\!\times\!d_2$ coefficients}
	\STATE Apply Radon transform perpendicular to corresponding v$^th$ orientation of the filter response (Resampled  $\psi_{(ABS-u,v)}$) and obtain GRG$_{u,v,j}$ ($\theta_\textrm{Radon}\leftarrow \theta_\textrm{Gabor}+90^\circ)$
		\STATE Get threshold: $T_{u,v,i}\!\leftarrow\!$ FindMedian(GRG$_{u,v,i}$)
		\STATE Binarize: $B_{u,v,i}\leftarrow$ Find(GRG$_{u,v,i} \geq T_{u,v,i}$)
		\STATE Append barcode: \\GRGBC$_{i}\!\leftarrow\!$ AppendRow (GRGBC$_{i},B_{u,v,i}$)
	\ENDFOR
\ENDFOR
\end{algorithmic}
\end{algorithm}                                                

\section{Experiments and Results}

\begin{figure}[htb]
\centering     
\subfigure[\tiny 1121-120-200-700]{\label{fig:b}\includegraphics[width=20mm,height=20mm]{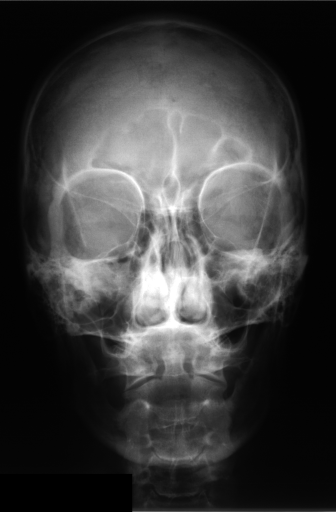}}
\subfigure[\tiny 1121-127-700-500]{\label{fig:a}\includegraphics[width=20mm,height=20mm]{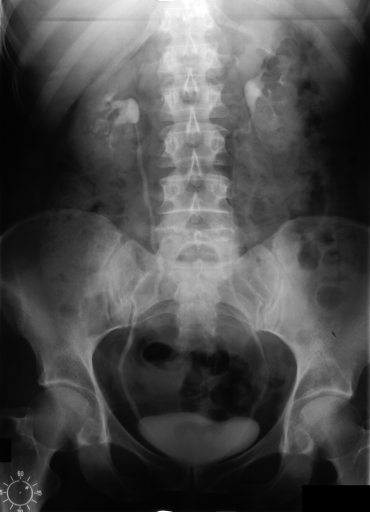}}
\subfigure[\tiny 1123-127-500-000]{\label{fig:b}\includegraphics[width=20mm,height=20mm]{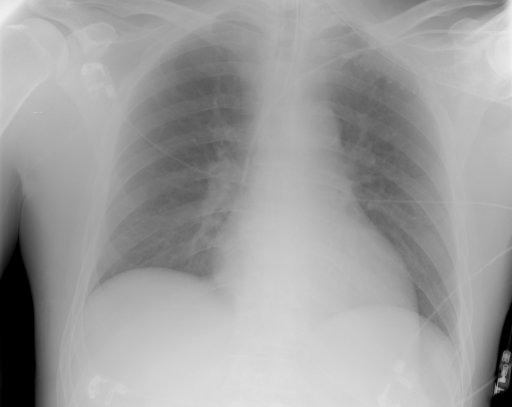}}
\caption{Sample x-rays with their IRMA codes.}
\label{fig:IRMASamples}
\end{figure}

\textbf{IRMA Dataset --} The Image Retrieval in Medical Applications (IRMA) database\footnote{http://irma-project.org/} is a collection of more than 14,000 x-ray images (radiographs) randomly collected from daily routine
work at the Department of Diagnostic Radiology of the RWTH Aachen University\footnote{http://www.rad.rwth-aachen.de/} \cite{H2,H3}. All images are classified into 193 categories (classes) and annotated with the IRMA code which relies on class-subclass relations to avoid ambiguities in textual classification. The IRMA code consists of four mono-hierarchical axes with three to four digits each: the technical code T (imaging modality), the directional code D (body orientations), the anatomical code A (the body region), and the biological code B (the biological system examined). The complete IRMA code subsequently exhibits a string of 13 characters, each in $\{0,\dots,9;a,\dots,z\}$:
TTTT-DDD-AAA-BBB.  More information on the IRMA database and code can be found in \cite{H2,H3}. IRMA dataset offers 12,677 images for training and 1,733 images for testing. Figure \ref{fig:IRMASamples} shows some sample images from the dataset long with their IRMA codes.
\vspace{-.2cm}

\textbf{Error Measurements --} As reported in literature on IRMA dataset \cite{H2,H3}, to evaluate the performance of the retrieval, an error evaluation score has been defined. Let an image have the IRMA code for its technical, directional, anatomical and biological axes. They can then be analyzed separately, summing the error over the individual axes \cite{H2,H3}. Let $l_1^I= l_1, l_2, \dots, l_i, \dots, l_I$ be the correct code (for one axis) of an image. As well, let $\hat{I}_1^I=\hat{l}_1, \hat{l}_2, \dots, \hat{l}_i, \dots, \hat{l}_I$ be the classified code (for one axis) of an image; where $l_i$ is specified precisely for every position, and in $\hat{l}i$ is allowed to say `don't know', which is encoded by `*'. Note that $I$ may be different for different images. Given a wrong classification at position $\hat{l}_i$ all succeeding decisions are considered to be wrong and, given a not-specified position, all succeeding decisions are considered to be not specified. Furthermore, no error is counted if the correct code is unspecified and the predicted code is a wildcard. In that case, all remaining positions are regarded as not specified. Wrong decisions that are easy (fewer possible choices at that node) are penalized over wrong decisions that are difficult (many possible choices at that node). A decision at position $l_i$ is correct by chance with a probability of $\frac{1}{b_i}$ if $b_i$ is the number of possible labels for position $i$. This assumes equal priors for each class at each position. Furthermore, wrong decisions at an early stage in the code (higher up in the hierarchy) are penalized more than wrong decisions at a later stage in the code (lower down on the hierarchy): i.e. $l_i$ is more important than $l_{i+1}$. Putting together: $\textrm{error} = \sum_{i=1}^{I} \frac{1}{b_i} \frac{1}{i} \delta(I_i,\hat{I}_i$  
with $\delta(I_i,\hat{I}_i)$ being 0, 0.5 or 1 for agreement, `don't know' and disagreement, respectively. 
The maximal possible error is calculated for every axis and the errors are normalized such that a completely wrong decision (i.e. all positions for that axis wrong) gets an error count of 0.25 and a completely correctly predicted axis has an error of 0. Thus, an image where all positions in all axes are wrong has an error count of 1, and an image where all positions in all axes are correct has an error count of 0. In addition to the error calculation scheme of imageCLEF, the code length $L_{\textrm{code}}$,  can be used to establish a suitability measure $\eta$ that prefers low error and short codes simultaneously:
\begin{equation}
\eta^k = \frac{\max\limits_i (E_{\textrm{total}}^i) \times \max\limits_i  (L_{\textrm{code}}^i)}{E_{\textrm{total}}^k \times L_{\textrm{code}}^k}
\end{equation}
Apparently, the larger $\eta$, the better the method, a desired quantification if the code length is  important in computation.

\textbf{Search --} The IRMA dataset consists of 193 classes, in which 12,677 IRMA images were used for training, and 1,733 images were used for testing. For each of the test images (that comes with its IRMA code) complete search was performed to find the most similar image whereas the similarity of an input image $I_i^\textrm{query}$ annotated with the corresponding barcode B$_i^\textrm{query}$ is calculated based on Hamming distance to any other image $I_j$  with its annotated barcode B$_j$:
\begin{equation}
\max_{j=1,2,3,\dots,1733; j\neq i} \left( 1 - \frac{| \textrm{XOR} (\textrm{B}_i^{\textrm{query}},\textrm{B}_j)|} {\textrm{B}_i^{\textrm{query}}} \right),
\end{equation}
where `B' can be RBC, GRGBC or GRIBC. The retrieval error $E_\textrm{total}$ of GRGBC and GRIBCs are provided in Table \ref{tab:results}. We carried our experiment with different window sizes \textcolor{black}{$s \times t=W\!\in \{11\!\times\! 11, 19\!\times\! 19, 23\!\times\! 23, 27\!\times\! 27\}$}. Based on our results, the filter window size of 23 pixels produces more discriminative Barcodes (more similarity in the Barcodes extracted from the images belonging to a certain class, and more contrast between the Barcodes extracted from images of two different classes), resulting in lower total error.

\begin{table*}[htb]
\centering
\caption{Comparing the performance of \textcolor{black}{(barcodes: $GRGBC(N_u,N_v,s,t)$), $ GRIBC(N_u,N_v,s,t)$, and RBC}, using two rankings based on $E_\textrm{total}$ and $\eta$. Every method indexed 12,677 images and was tested with 1,733 images from the IRMA dataset.}
\label{tab:results}
\begin{tabular}{llll||lllll}
Barcode            & $E_\textrm{total}$   & $L_\textrm{code}$ & Time & Barcode            & $E_\textrm{total}$   & $L_\textrm{code}$ & Time & $\eta$ \\ \hline\hline
GRGBC(8,16,23,23)  & \CL 322.41 & 6282  & 0.345     & RBC4               & 476.62   & 512   & 0.007     & \CL 16.85  \\
GRGBC(8,20,23,23)  & \CL 322.98  & 7840  & 0.432     & GRGBC(5,4,23,23)   & 441.78 & 980   & 0.060     & \CL 9.49 \\
GRGBC(10,16,23,23) & \CL 326.95 & 7840  & 0.430       & RBC8               & 478.54   & 1024  & 0.007     & \CL 8.39 \\
GRIBC(5,16,23,23)  & \CL 330.36 & 5120  & 0.201     & GRIBC(5,4,23,23)   & 417.24 & 1280  & 0.080     & \CL 7.69 \\
GRIBC(5,20,23,23)  & \CL 332.72  & 6400  & 0.244     & GRGBC(5,8,23,23)   & 381.67 & 1960  & 0.112     & \CL 5.49  \\
GRIBC(4,20,23,23)  & \CL 338.09 & 5120  & 0.202     & GRIBC(5,8,23,23)   & 338.75 & 2560  & 0.119     & \CL 4.74 \\
GRIBC(8,16,23,23)  & \CL 338.49 & 8192  & 0.301     & RBC16              & 470.57   & 2048  & 0.007     & \CL 4.26 \\
GRIBC(5,8,23,23)   & \CL 338.75 & 2560  & 0.119     & GRGBC(5,12,23,23)  & 380.15 & 2940  & 0.165     & \CL 3.67 \\
GRIBC(7,12,23,23)  & \CL 339.66 & 5376  & 0.209     & GRGBC(5,16,23,23)  & 365.91 & 3920  & 0.219     & \CL 2.86 \\
GRIBC(5,16,17,17)  & \CL 339.05 & 5120  & 0.165     & GRIBC(5,16,23,23)  & 330.36 & 5120  & 0.201     & \CL 2.43 \\
GRGBC(5,16,23,23)  & \CL 365.91 & 3920  & 0.219     & GRIBC(4,20,23,23)  & 338.09 & 5120  & 0.202     & \CL 2.37 \\
GRGBC(5,20,23,23)  & \CL 367.34   & 4900  & 0.326     & GRIBC(5,16,17,17)  & 339.05 & 5120  & 0.165     & \CL 2.36 \\
GRGBC(5,12,23,23)  & \CL 380.15 & 2940  & 0.165     & GRGBC(5,20,23,23)  & 367.34   & 4900  & 0.326     & \CL 2.28 \\
GRGBC(5,8,23,23)   & \CL 381.67 & 1960  & 0.112     & GRIBC(7,12,23,23)  & 339.66 & 5376  & 0.209     & \CL 2.25 \\
GRIBC(5,4,23,23)   & \CL 417.24 & 1280  & 0.080     & RBC32              & 475.92   & 4096  & 0.008     & \CL 2.10 \\
GRGBC(5,4,23,23)   & \CL 441.78 & 980   & 0.060     & GRGBC(8,16,23,23)  & 322.41 & 6282  & 0.345     & \CL 2.03 \\
RBC16              & \CL 470.57   & 2048  & 0.007           & GRIBC(5,20,23,23)  & 332.72  & 6400  & 0.244     & \CL 1.93 \\
RBC32              & \CL 475.92   & 4096  & 0.008     & GRGBC(8,20,23,23)  & 322.98  & 7840  & 0.432     & \CL 1.62 \\
RBC4               & \CL 476.62   & 512   & 0.007    & GRGBC(10,16,23,23) & 326.95 & 7840  & 0.430       & \CL 1.60 \\
RBC8               & \CL 478.54   & 1024  & 0.007     & GRIBC(8,16,23,23)  & 338.49 & 8192  & 0.301     & \CL 1.48 \\
\end{tabular}
\end{table*}

\textbf{Ranking based on Error --}  As apparent from Table \ref{tab:results} (left section), the proposed combinations of Radon and Gabor features to generate more expressive barcodes  clearly perform better compared to RBC, LBP and LRBP as validated in literature \cite{3}. The lowest total error of 322.41 for GRGBC(8,16,23,23) is roughly equal a retrieval accuracy of  $81\%$ assuming that based on IRMA error calculation, if all digits are wrong, then we received 1 for that query result. Since we have 1,733 test images, one may calculate the accuracy $A$ as $A=1-\frac{E_\textrm{total}}{1733}$. one has to bear in mind that the error, or the approximate accuracy, is calculate for the first hit only. In literature, the error or accuracy is determined for the top three or top five results, which can lower the total error. 

\textbf{Ranking based on Error and Code Length -- }  Table \ref{tab:results} (right section) shows that when we simultaneously consider error and code length (e.g., via the suitability measure), the ranking of barcode methods drastically changes. Here, RBC with 4 projections angles takes the first spot although it has an total error score of 476.62 (equivalent to $72\%$ accuracy). When we examine the suitability numbers $\eta$ and compare them against the respective time measurements for each type of barcode, they show a consistent direct relationship. Of course, longer barcodes will prolong the retrieval time.

\section{Summary and Conclusions}
Content-based image retrieval can benefit from introduction of robust and compact binary features. Barcodes seem to be a promising candidate in this regard. Radon barcodes, recently introduced, employ a simple thresholding of Radon projections to generate a barcode that can be used to tag images for faster and more accurate retrieval tasks. In this paper, we combined Radon and Gabor transform in two different ways to generate new types of barcodes. We focused on binarized versions of Gabor-of-Radon-Image (GRI) and Guided-Radon-of-Gabor (GRG) features, to not only save memory space for the task of archiving but also to expedite the search when applied on large archives. We employed the IRMA dataset with 14,4100 images to validate the performance to the proposed barcodes. The application of barcodes is not limited to the medical x-ray images. They can be employed as powerful methods to extract discriminative features from all types of images. In \textcolor{black}{our} future works, we attempt to validate the barcodes on other publicly available datasets. 

\bibliographystyle{abbrv}

\end{document}